\documentclass[11pt]{article}
\usepackage{amsfonts}
\usepackage{amsmath}
\usepackage{amsthm}
\usepackage{algorithm}
\usepackage{algpseudocode}
\usepackage[preprint]{acl}
\usepackage{multirow}
\usepackage{booktabs} 
\usepackage{times}
\usepackage{latexsym}

\usepackage[T1]{fontenc}
\usepackage{bbm}
\usepackage[utf8]{inputenc}

\usepackage{microtype}

\usepackage{inconsolata}

\usepackage{graphicx}
\usepackage{xcolor}

\title{Planner-Guided Recursive Exploration \\for Diffusion Language Model Decoding}

\author{\\
\textbf{~~ Yingqian Cui$^{1}$~~ Wei Deng$^{2}$ ~~ Lantao Mei~~ Hang Li$^{1}$} \\
\textbf{Charu C. Aggarwal$^{3}$ ~~ Hui Liu$^{1}$~~ Yue Xing$^{1}$ }\\
 ~~ $^{1}$Michigan State University
 ~~$^{2}$ Morgan Stanley ~~  $^{3}$ IBM T.J. Watson Research Center  \\}

\begin{document}
\maketitle
\begin{abstract}
Masked diffusion language models (DLMs) enable parallel text generation by iteratively refining masked tokens, offering a promising alternative to autoregressive decoding.
Recent search-based decoding methods improve the accuracy--efficiency trade-off by exploring multiple candidate decoding trajectories before committing token updates.
However, existing search-based decoding methods face two key limitations. First, they offer limited guidance on when search is actually beneficial for the current decoding state. Second, they typically rely on a shallow one-step search horizon, which may favor locally promising but suboptimal decoding trajectories.
To address these limitations, we propose \textbf{PRED}, a planner-guided recursive exploration framework for DLM decoding.
PRED employs a learned planner to determine when search should be triggered, and performs adaptive multi-step rollout to evaluate candidate trajectories over longer horizons.
Moreover, the same planner is recursively applied to intermediate rollout states to determine whether additional branch expansion is needed, allowing PRED to re-initiate search 
when necessary.
Experiments across multiple benchmarks and DLM backbones show that PRED achieves a better accuracy--efficiency trade-off than existing confidence-based and search-based decoding methods.

\end{abstract}

\section{Introduction}
\vspace{-0.02in}
\label{sec:intro}
\vspace{-0.03in}

Masked diffusion language models (DLMs) have recently emerged as an alternative to traditional autoregressive (AR) models for text generation.
Unlike AR models, which generate tokens sequentially in a token-by-token manner, DLMs operate by iteratively refining a sequence of masked tokens, enabling parallel decoding across multiple positions~\cite{sahoo2024simple,nie2026large,ye2025dream}.
This property offers a fundamental advantage over AR decoding:
By relaxing the strict left-to-right dependency, DLMs can potentially improve generation efficiency and enable more flexible inference strategies.

%
{While the diffusion framework supports parallel token updates in principle, its denoising-based training objective does not prescribe a specific decoding order or token selection strategy, leaving a large design space for inference-time decoding that can substantially affect generation quality and efficiency.}
To better exploit the parallel nature of DLMs, a growing line of work focuses on designing efficient decoding strategies. 
Existing methods have evolved from fixed-step parallel decoding~\cite{nie2026large} to more adaptive confidence-aware decoding frameworks~\cite{yu2025dimple,wu2026fast,wei2026accelerating}.  

More recently, search-based decoding strategies have begun to explore multiple possible token commitments before making a decoding decision. For example, \citet{fu2025bits, lee2025lookahead} proposed to expand several candidate token updates and evaluate their downstream decoding states before selecting a branch to commit. Such methods can be viewed as performing a lightweight search over partially decoded sequences, allowing the decoder to compare alternative trajectories rather than relying solely on local confidence estimates. 

Despite their effectiveness, existing search-based decoding methods still face two major limitations. \textbf{First, existing approaches offer limited guidance on when search is actually beneficial.}
Existing approaches typically either perform search at every decoding step or rely on hand-crafted confidence-based rules to determine when search should be triggered. While always invoking search ensures sufficient exploration, it incurs substantial computational overhead by allocating search effort even to states that can already be decoded reliably without additional exploration. Confidence-based triggering is more selective and reduces unnecessary computation, but it relies on coarse uncertainty signals that do not directly indicate whether additional exploration will improve the downstream decoding trajectory. As a result, neither strategy provides an ideal mechanism for allocating search computation across heterogeneous decoding states.


Second, \textbf{existing search is typically limited to a short, one-step horizon}. Candidate token updates are evaluated primarily according to their immediate effect on the next decoding state. For example, \citet{fu2025bits} propose to measure how many additional high-confidence predictions are induced after a single rollout step. While this provides a useful local signal, it does not necessarily reflect the quality of the resulting longer-term decoding trajectory.
From a longer-horizon perspective, such local exploration can be suboptimal: the token that maximizes immediate confidence gain may not lead to the best future trajectory, causing the search to be trapped in a locally favorable but globally suboptimal commitment.

{Addressing these two limitations requires more than straightforward
extensions of existing search strategies. For search triggering, a natural
solution is to replace hand-crafted confidence rules with a learned decision
mechanism. However, training such a mechanism is challenging because there is
no direct supervision indicating whether search should be triggered at a
particular decoding state. For longer-horizon search, although increasing the rollout depth may intuitively appear beneficial, our initial experiments show that naively extending the rollout linearly to more steps can instead degrade performance (related results can be found in Section~\ref{sec:nai}). This is because the required exploration depth varies across decoding states, while a fixed linear rollout cannot adapt when intermediate states require renewed branching.}

To address these challenges, we propose \textbf{PRED}, \textbf{P}lanner-guided \textbf{R}ecursive \textbf{E}xploration for \textbf{D}iffusion language model Decoding, an adaptive search framework that dynamically controls both \textbf{when to search} and \textbf{how far to search} in decoding.

Specifically, PRED has three key components:

{(1) \textbf{Planner-guided search triggering}. To determine {when to trigger search} at the current decoding state, PRED employs a lightweight planner that predicts whether additional exploration would be beneficial. To develop the planner, we first initialize it using search decisions collected from successful confidence-guided decoding trajectories, and then refine it with on-policy trajectory-level reinforcement learning based on downstream generation quality and decoding cost.

(2) \textbf{Adaptive multi-step exploration}. To flexibly determine how far the search should proceed, PRED evaluates candidate trajectories over multiple rollout steps and adaptively adjusts the rollout depth according to how clearly the candidate trajectories can be distinguished.

(3) \textbf{Recursive branch expansion}.
To further ensure the effectiveness of the multi-step exploration, PRED explicitly considers whether new branch expansion is needed at intermediate states. Specifically, PRED repeatedly applies the same planner throughout the rollout process to identify intermediate states that require further expansion, allowing new rounds of search to be selectively initiated.}


Together, these components enable PRED to perform adaptive and recursive search throughout the decoding process, providing deeper exploration when needed while avoiding unnecessary rollout.

To validate the effectiveness of the proposed method, we conduct experiments on multiple benchmarks using different DLM backbones. Across a range of decoding budgets, PRED consistently achieves a better accuracy--decoding step trade-off than existing confidence-based and search-based decoding methods.

\vspace{-0.03in}
\section{Related Works}
\vspace{-0.03in}
\noindent\textbf{Diffusion Language Models.}
Diffusion models have achieved remarkable success in continuous domains such as image and audio generation~\cite{ho2020denoising,song2020score,kong2020diffwave}, which has motivated efforts to extend the diffusion paradigm to discrete text. 
Early work on discrete diffusion, notably D3PM~\cite{austin2021structured}, established a general framework for defining noising processes over categorical variables. Subsequent works leverage this masking-based formulation to develop masked diffusion language models (MDLMs) that generate text via iterative unmasking~\cite{sahoo2024simple,shi2024simplified,ou2025your}.
Building on these foundations, recent large-scale MDLMs have reached performance competitive with autoregressive counterparts: LLaDA~\cite{nie2026large} trains an 8B model from scratch with bidirectional attention, Dream~\cite{ye2025dream} develops a diffusion-based LLM initializing from a pretrained autoregressive model, and LLaDA 2.0~\cite{bie2025llada2} further scales to 100B parameters with a Mixture-of-Experts architecture.

\noindent\textbf{Decoding Strategies for DLMs.}
To fully leverage the parallel nature of DLMs, a growing line of work focuses on designing efficient decoding strategies.
Initial attempts focus on fixed-step parallel decoding~\cite{nie2026large,ye2025dream}, while subsequent works introduce more adaptive confidence-aware decoding strategies, including fixed and dynamic threshold decoding~\cite{yu2025dimple,wu2026fast}.
Other studies further improve decoding efficiency through multi-stage decoding strategies~\cite{wei2026accelerating}, confidence calibration~\cite{huang2025pc}, and entropy-bounded unmasking~\cite{ben2026accelerated}. In addition, \citet{chen2025dultra} introduce a learned planner head trained via reinforcement learning to guide token unmasking decisions during decoding. More recently, parallel search-based decoding methods explore multiple candidate decoding trajectories before committing token updates~\cite{lee2025lookahead,fu2025bits}. 
However, as discussed in Section~\ref{sec:intro}, existing methods lack effective mechanisms for determining when search is beneficial and typically restrict exploration to a shallow, one-step horizon, leading to suboptimal search decisions and limited long-horizon planning capabilities. 
Notably, our work shares a similar emphasis with \citet{fu2025bits} on improving the accuracy--decoding efficiency trade-off, whereas \citet{lee2025lookahead} primarily aims to maximize accuracy without explicitly considering the number of decoding steps.

\vspace{-0.03in}
\section{Preliminary}
\vspace{-0.05in}

In this section, we introduce the diffusion decoding setup and the fast block diffusion sampling method~\cite{fu2025bits} that is used as the decoding backbone of our framework.



\noindent\textbf{Diffusion Decoding Procedure.} Masked diffusion language models generate a sequence through iterative denoising. Let $x_t$ denote the partially decoded sequence at decoding step $t$, and let $\mathcal{M}_t$ denote the set of masked positions in $x_t$. Given $x_t$, the model predicts a token distribution for each masked position $i \in \mathcal{M}_t$, with prediction confidence denoted as $c_t^i(x_t)$. At each decoding step, a subset of masked positions satisfying $c_t^i(x_t) > C$ is selected and committed according to their predicted tokens, producing the next partially decoded state.

\noindent\textbf{Fast Block Diffusion Decoding.} Block diffusion decoding generates the sequence in a block-by-block manner, where a sequence of length $n$ is partitioned into $L$ decoding blocks with block size $n_b$. 
Within each block, the decoder performs iterative refinement before proceeding to the next block~\cite{nie2026large}.
Fast block sampling further improves decoding efficiency by assigning a fixed budget of $N$ decoding steps to each block and unlocking the next block once the budget is exhausted \citep{fu2025bits}. As a result, earlier blocks may still contain masked tokens when later blocks become available. The decoder can commit high-confidence tokens across all unlocked blocks in parallel, increasing the number of tokens decoded at each step.

\vspace{-0.05in}
\section{PRED: Planner-Guided Recursive Exploration}
\vspace{-0.05in}
\label{sec:pred}
In this section, we present PRED, our planner-guided recursive exploration framework for diffusion language model decoding. 
In Section~\ref{sec:framework}, we first introduce the overall inference workflow of PRED and describe how search is organized throughout the decoding process. We then further describe the formulations for candidate construction in Section~\ref{sec:candidate}.  Section~\ref{sec:planner} introduces the search planner used in PRED, including its architecture and two-stage training procedure. Finally, Section~\ref{sec:hyperparameter} discusses the hyperparameter selection.

\vspace{-0.05in}
\subsection{PRED Inference Framework}\label{sec:framework}
\vspace{-0.05in}

\begin{figure*}
    \centering
     \vspace{-0.07in}
    \includegraphics[width=0.93\linewidth]{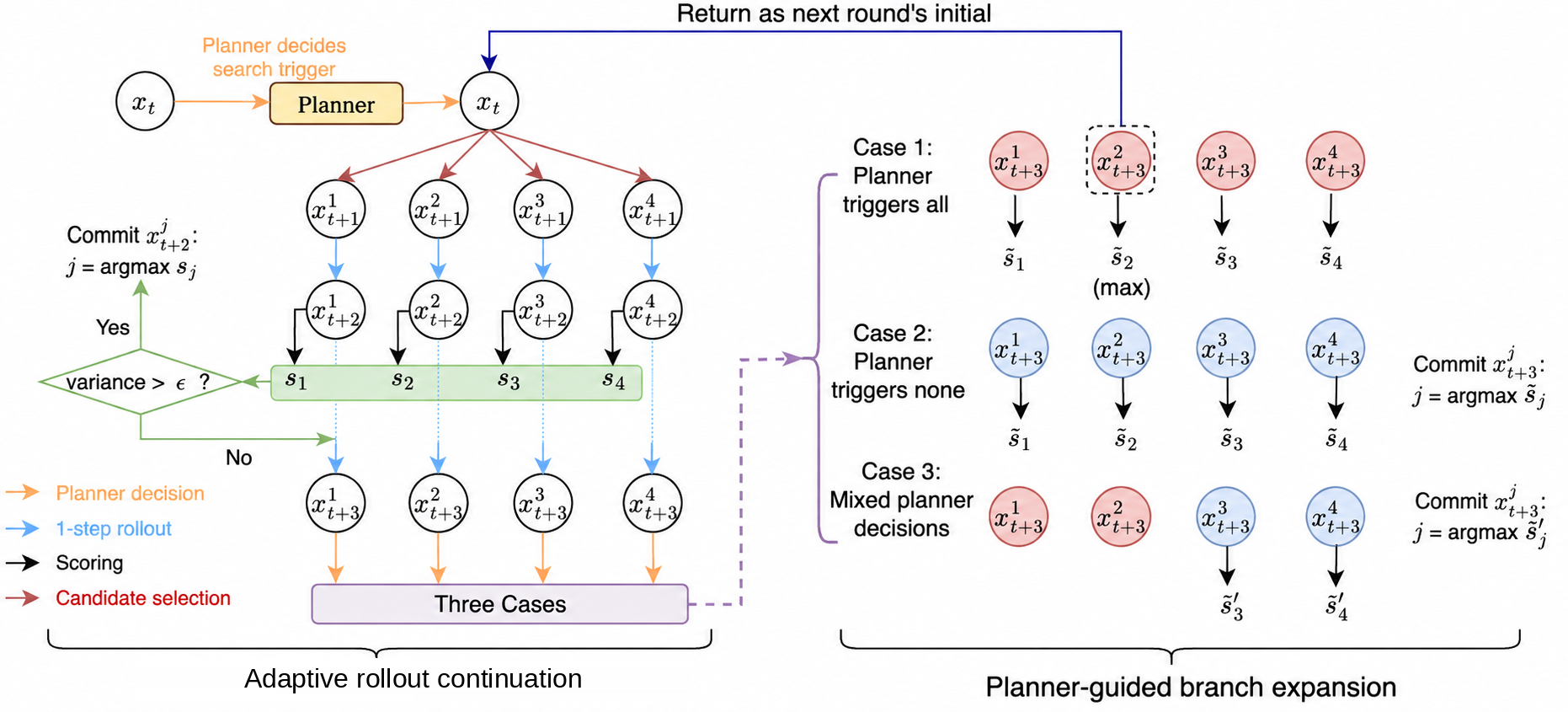}
    \vspace{-0.12in}
    \caption{\small Overview of the planner-guided recursive exploration framework (when the maximum rollout depth is 2).}
    \label{fig:flow}
    \vspace{-0.2in}
\end{figure*}

{\noindent\textbf{Framework Overview.} The overall workflow is illustrated in Figure~\ref{fig:flow}. For simplicity, the figure illustrates a special case with the maximum rollout depth set to 2. The detailed algorithm of PRED is shown in Algorithm~\ref{alg:ours} in Appendix~\ref{appd:alg}. }{As listed in Section \ref{sec:intro}, PRED contains three steps: (1) planner-guided search triggering, (2) adaptive multi-step exploration, and (3) recursive branch expansion.}

\noindent\textbf{Inference Workflow.}
As illustrated in Figure~\ref{fig:flow}, given a partially decoded state $x_t$, PRED first applies a lightweight search planner to determine whether additional exploration should be triggered. If search is not triggered, decoding proceeds using the standard confidence-threshold decoding procedure. Otherwise, PRED selects a set of $k$ candidate positions $\mathcal{H}_t$ and constructs a candidate branch for each $j\in\mathcal{H}_t$ by tentatively committing the predicted token at position $j$. The resulting candidate states $\{x_t^{(j)}\}_{j\in\mathcal{H}_t}$ are then evaluated in parallel through forward rollout.

Once search is initiated, PRED adaptively determines how far the candidate trajectories should be explored. At each rollout depth $r$, each candidate branch $j \in \mathcal{H}_t$ is assigned a cumulative trajectory score $\tilde{s}^{(r)}(j)$ to quantify its downstream decoding benefit (whose detailed formulation is provided in Section~\ref{sec:candidate}). PRED measures the distinguishability among the candidate trajectories using the score variance $\mathrm{Var}(\{\tilde{s}^{(r)}(j)\}_{j=1}^{k})$. When the variance remains below a predefined threshold $\tau$, indicating that the candidate trajectories are not yet sufficiently distinguishable, PRED continues the rollout to obtain additional future information. Once the variance exceeds $\tau$, or the maximum rollout depth $T_{\max}$ is reached, PRED terminates the current rollout and selects the highest-scoring trajectory.

Meanwhile, PRED repeatedly applies the same planner to the intermediate state of each candidate branch to determine whether a new round of branch expansion is needed. {As shown in the right part of Figure~\ref{fig:flow},} based on the expansion status of the candidate branches, PRED considers three cases.

\noindent\textbf{(1) No branch requires further expansion.}
PRED retains all candidate trajectories in $\mathcal{H}$ and continues the adaptive rollout procedure. After each additional rollout step, the trajectory scores $\{\tilde{s}^{(r)}(j)\}_{j\in\mathcal{H}}$ are updated, and the rollout continues until their variance exceeds $\tau$ or the maximum rollout depth $T_{\max}$ is reached.

\noindent\textbf{(2) All branches require further expansion.}
Rather than expanding all candidate trajectories and rapidly increasing the search space, PRED selects the most promising intermediate state according to both its accumulated trajectory score and the confidence of its current decoding frontier (with the detailed formulation provided in Section~\ref{sec:candidate}). The selected state is then used as the starting point for a new search round, from which a new set of candidate branches is constructed.

\noindent\textbf{(3) Only a subset of branches requires further expansion.}
PRED removes the branches that require additional expansion from $\mathcal{H}$ and retains the remaining trajectories. The adaptive rollout procedure then continues over the reduced candidate set until a single branch remains, the score variance exceeds $\tau$, or the rollout budget $T_{\max}$ is exhausted.

When the adaptive rollout terminates in Cases~(1) or~(3), PRED selects the highest-scoring remaining trajectory and commits it to the main decoding process. In Case~(2), the selected intermediate state instead serves as the root of a new search round, and the same procedure is recursively repeated.

Through this process, PRED jointly adapts the depth and structure of search: trajectory-score variance controls how far the current candidate branches are rolled out, while the planner determines whether newly reached states require additional branch expansion. We next describe the candidate construction and hypothesis evaluation procedures used within each search round.

\vspace{-0.04in}
\subsection{Candidate Construction and Trajectory Evaluation}\label{sec:candidate}
\vspace{-0.03in}

\paragraph{Candidate Construction} 
Following the strategy of \citet{fu2025bits}, once exploration is triggered, the method identifies a set of informative candidate positions from the masked positions:

{\vspace{-0.2in}\small
\begin{equation*}
    \mathcal{H} = \mathrm{Topk}_{i \in \mathcal{M}_t}
\left( - \left| c_t^i(\mathbf{x}_t) - c^{\mathrm{info}} \right|
+ \beta \cdot (i-(b_t-1)n_b) \right),
\end{equation*}}
\vspace{-0.25in}

\noindent where $\mathbf{x}_t$ denotes the partially decoded sequence at decoding step $t$, with masked positions $\mathcal{M}_t$. $c_t^i(\mathbf{x}_t)$ denotes the confidence at position $i$, $b_t$ is the index of the current decoding block at step $t$, $n_b$ is the block size, and $\beta$ is used to balance the two terms.

The first term selects positions whose confidence is close to the target exploration level $c^{\mathrm{info}}$, which captures tokens that are uncertain but potentially informative.  Following the empirical findings of \citet{fu2025bits}, we set $c^{\mathrm{info}}=0.2$.
{The second term uses \(i-(b_t-1)n_b\), the relative position of token \(i\) within the current block, to assign slightly larger scores to later positions when \(\beta>0\), thereby encouraging the decoding frontier to move forward.}

\paragraph{Trajectory Scoring.}
For each candidate position \(j \in \mathcal{H}\), when the rollout depth is \(r\), PRED obtains the corresponding intermediate state \(x_{t+r;j}\) and evaluates the candidate trajectory as:
\vspace{-0.12in}
\begin{equation}
{\small
\begin{aligned}
\tilde{s}^{(r)}(j)= \alpha \log c_t^j(\mathbf{x}_t)
+ \log ( \sum_{\ell=1}^{r} \sum_{i \in \mathcal{T}_j^\ell} c_{t+\ell}^i(\mathbf{x}_{t+\ell;j})).
\end{aligned}}\label{eq:s}
\end{equation}
\vspace{-0.18in}

\noindent Here, $\mathcal{T}_j^\ell$ denotes the set of positions that become high-confidence and are selected for commitment at the $\ell$-th rollout step along candidate trajectory $j$. The first term measures the confidence of the initially explored token, while the second term accumulates the confidence of downstream committed positions induced by the candidate over all rollout steps up to depth $r$. $\alpha$ controls the relative contribution of the initial candidate confidence.


Intuitively, $\tilde{s}^{(r)}(j)$ captures both the reliability of the initial token commitment and its longer-horizon decoding effect. A higher score indicates that committing $j$ leads to a trajectory that unlocks more high-confidence predictions during rollout.

\paragraph{Recursive Search Root Selection.}
When all candidate branches require further expansion, PRED selects one intermediate state as the root of the next search round to avoid expanding all branches simultaneously. Specifically, the selected branch is
\vspace{-0.12in}
\begin{equation*}
j^*=\arg\max_{j\in\mathcal{H}} \left(\tilde{s}^{(r)}(j)+\log \bar{c}_{F}^{j}\right),
\end{equation*}
\vspace{-0.2in}

\noindent where $\bar{c}_{F}^{j}$ denotes the average confidence over the decoding frontier of the intermediate state along trajectory $j$. This criterion jointly considers the accumulated downstream decoding benefit of the trajectory and the confidence of its current intermediate state. The selected state $x_{t+r;j^*}$ is then used as the root of the next search round.




\vspace{-0.02 in}
\subsection{Learning the Search Planner}
\vspace{-0.02in}
\label{sec:planner}

\paragraph{Planner architecture.}
Given a partially decoded state $x_t$, let $\{(\mathbf{h}_i,c_i,p_i)\}_{i\in\mathcal{M}_t}$ denote the hidden representation, prediction confidence, and within-block position of each remaining masked token, respectively. Let $\mathbf{q}_t$ denote the decoding configuration, including the confidence threshold $C$ and per-block decoding budget $N$. The planner is parameterized by $\phi$ and computes
\vspace{-0.05in}
\begin{equation*}
\ell_t=f_\phi\left( \{(\mathbf{h}_i,c_i,p_i)\}_{i\in\mathcal{M}_t}, \mathbf{q}_t
\right),
\end{equation*}
\vspace{-0.22in}

\noindent where $\ell_t$ is a scalar logit indicating whether additional search should be triggered.

The planner $f_\phi$ projects the token-level inputs into a shared representation space and adds a learnable positional embedding based on each token's within-block position. The resulting token features are processed by a lightweight Transformer block and aggregated through masked mean pooling to obtain a state-level representation. This representation is then concatenated with the decoding configuration $\mathbf{q}_t$ and passed through an MLP head to produce $\ell_t$. This design allows the planner to jointly capture interactions among masked positions and the global decoding condition when deciding whether additional search is beneficial.

\paragraph{Training Procedure} The search planner is trained in two stages. We first use a confidence-based triggering policy to provide supervised warm-up signals, and then refine the planner with trajectory-level reinforcement learning.

\noindent\textit{(1) Supervised warm-up.}
We initialize the planner through supervised training using search decisions generated by the confidence-based triggering strategy of \citet{fu2025bits}.
Specifically, we follow the same decoding workflow described in Section~\ref{sec:framework}, but replace the learned planner with a rule-based confidence trigger. Given a decoding state $x_t$, search is triggered when the average confidence over the current decoding frontier $\mathcal{F}_t$ falls below a threshold $\gamma$ and the number of remaining masked positions exceeds $N_e$:
\vspace{-0.06 in}
\begin{equation*}
y_t =
\mathbb{I}
\left[
\bar{c}_{\mathcal{F}_t} < \gamma
\;\land\;
|\mathcal{M}_t| > N_e
\right],
\end{equation*}
\vspace{-0.25in}

\noindent where $y_t=1$ indicates triggering search and $y_t=0$ indicates proceeding without additional exploration. 
During generation, we collect the encountered decoding states and extract the corresponding token-level features and decoding configuration described in the planner architecture, together with their search decisions. We retain state-decision pairs only from trajectories that produce a correct final answer, so that the supervision is derived from successful decoding behavior. Each retained state is treated as a binary classification example, with the corresponding rule-based decision serving as the training label. The planner is then trained with a binary cross-entropy objective to predict these search decisions. This supervised stage provides a stable initialization before further policy refinement with reinforcement learning.

\noindent\textit{(2) GRPO refinement.}
After supervised warm-up, we further refine the planner using a trajectory-level GRPO objective~\cite{shao2024deepseekmath}. For each prompt, we generate a group of decoding trajectories by repeatedly executing the PRED decoding workflow with the current planner parameters $\phi$. At each planner invocation, a binary action is sampled according to the search probability predicted by the planner.  
Each trajectory $i$ receives a reward based on answer correctness and
decoding cost:
\vspace{-0.09in}
\begin{equation*}
R_i = \mathbb{I}[\mathrm{correct}_i]\left(
\lambda_{\mathrm{acc}} - \lambda_{\mathrm{cost}} S_i \right),
\end{equation*}
\vspace{-0.21in}

\noindent where $S_i$ denotes the number of model forward steps used by trajectory $i$. $\lambda_{\mathrm{acc}}$ controls the reward assigned to a correct trajectory, while $\lambda_{\mathrm{cost}}$ controls the penalty on
decoding cost. With this formula, incorrect trajectories receive zero reward, while among correct trajectories, those requiring fewer decoding steps receive higher rewards. For each trajectory, a group-relative advantage $A_i$ is computed by comparing its reward with those of other trajectories generated from the same prompt. The same trajectory-level advantage is shared by all planner decisions along that trajectory.

The policy objective is formulated as

{\small \vspace{-0.2in} \begin{equation*}
\mathcal{L}_{\mathrm{policy}}
= -\frac{1}{G} \sum_{i=1}^{G} A_i \sum_{t\in\mathcal{T}_i} \log \pi_\phi(a_{i,t}\mid s_{i,t}),
\end{equation*}}
\vspace{-0.18in}

\noindent where $G$ is the group size, $\mathcal{T}_i$ denotes the set of planner decisions along trajectory $i$, and $a_{i,t}$ denotes the sampled search decision at state $s_{i,t}$. $\pi_\phi(a_{i,t}\mid s_{i,t})$ denotes the Bernoulli policy induced by the planner output at state $s_{i,t}$. We additionally regularize the planner toward the frozen supervised warm-up policy using a KL penalty, with an optional entropy bonus to encourage exploration.

\vspace{-0.04in}
\subsection{Hyperparameter Selection}\label{sec:hyperparameter}
\vspace{-0.04in}

Our method introduces several hyperparameters that control candidate selection and decoding progression.  
We discuss our hyperparameter selection strategy in this subsection.
Specifically, we group the hyperparameters into two main categories:

\noindent\textbf{1. Fixed hyperparameters }
We fix $k=4$, $\alpha=0.1$, $\beta=0.01$, $T_\text{max}=3$, $c_{\mathrm{info}}=0.2$ and $\tau=0.01$ in all experiments. 
These values are either adopted from prior work~\citep{fu2025bits} or selected by calibration. 
We observe no significant dataset-specific variation for these hyperparameters.


\noindent\textbf{2. Calibration-guided hyperparameters }
For hyperparameters that are not fixed, we use a small calibration set to evaluate different combinations and identify selection rules. Since parameters $C$ and $N$ induce different accuracy-efficiency trade-offs, we analyze the configurations on the Pareto frontier to identify the best parameter patterns.

In particular, we observe a clear interaction between the confidence threshold $C$ and the per-block unlocking budget $N$. Both parameters affect the decoding pace: smaller $N$ unlocks subsequent blocks earlier, while smaller $C$ commits more tokens per step. Both choices can reduce the number of decoding steps, but may also make the decoding process more aggressive.
Among Pareto-frontier configurations, the preferred threshold tends to increase slowly with the decoding budget. This empirical trend can be roughly summarized as:

{ \small \vspace{-0.15in}
\begin{equation*}
 C^*(N) \in  \left[ 0.5 + 0.05\log_2 N,\; 0.5 + 0.10\log_2 N \right].
\end{equation*}}
\vspace{-0.25 in}


To explain this relationship, when $N$ is small, the decoding frontier advances more quickly because subsequent blocks are unlocked after fewer decoding rounds. 
In this case, a lower confidence threshold is often preferred so that the decoder can commit enough reliable tokens before the frontier moves forward, providing useful context for later blocks. 
In contrast, when $N$ is large, the decoder spends more refinement rounds on the current block before unlocking later blocks. {We provide justification about the log-scale relationship in Appendix~\ref{appd:theo}.}
Notably, these trends are generally consistent across different datasets, suggesting that the calibration rules are not strongly dataset-specific.

\vspace{-0.05in}
\section{Empirical Evaluation} \label{sec:exp}
\vspace{-0.05in}
\subsection{Experimental Setups}\label{sec:setup}
\vspace{-0.02in}
\textbf{Models, Benchmarks and Baselines} We conduct experiments using two base DLMs, \textbf{LLaDA-8B-Instruct}~\cite{nie2026large} and \textbf{Dream-v0-Instruct-7B}~\cite{ye2025dream}. 
We evaluate PRED on four widely used benchmarks, including \textbf{HumanEval}~\cite{chen2021evaluating}, \textbf{GSM8K}~\cite{cobbe2021training}, \textbf{MATH500}~\cite{hendrycks2021measuring}, and \textbf{BBH}~\cite{suzgun2023challenging}, covering mathematical reasoning, code generation, and general reasoning tasks. For HumanEval, GSM8K, and MATH500, we use the full evaluation sets, containing 164, 1319, and 500 examples, respectively. For BBH, we randomly sample 1000 examples from the test set for evaluation to keep the evaluation cost manageable.
For all benchmarks, we use a generation length of 256 and a block size of 32. All experiments are performed using greedy decoding and run on NVIDIA H200 GPUs. 
We consider four baselines: (1) Conf, the basic confidence-threshold decoding strategy; (2) Fast-dLLM~\cite{wu2026fast}, an adaptive confidence-aware decoding method that adjusts token commitment beyond a fixed threshold; {(3) ETE~\cite{fu2025bits}, a search-based decoding method with confidence-based triggering and one-step rollout; and (4) LookUM~\cite{lee2025lookahead}, a search-based decoding method with always-on triggering and one-step rollout.}

\noindent\textbf{Planner Training Details.}
For each base model, we use a single search planner trained on the combined training data from GSM8K and MATH, and apply it across all evaluation benchmarks for that model. Both the supervised warm-up and GRPO refinement stages are conducted using this combined training set. More details about the training hyperparameters are provided in Appendix~\ref{appd:training}.

\noindent\textbf{Hyperparameter Configurations.}
Both ETE and PRED involve several hyperparameters. 
As discussed in Section~\ref{sec:hyperparameter}, the hyperparameters that need to be varied across configurations for both methods are the per-block unlocking budget \(N\) and  confidence threshold \(C\). 
We follow the calibration guideline in Section~\ref{sec:hyperparameter}. 
Specifically, we choose \(N\) from powers of two in the range \([1,32)\), and pair each \(N\) with confidence thresholds \(C\) suggested by the corresponding range-valued rule, forming a set of accuracy--efficiency trade-off configurations. 




\vspace{-0.06in}
\subsection{Main Results}\label{sec:main}
\vspace{-0.05in}

\begin{figure*}
\vspace{-0.05in}
    \centering
    \includegraphics[width=0.93\linewidth]{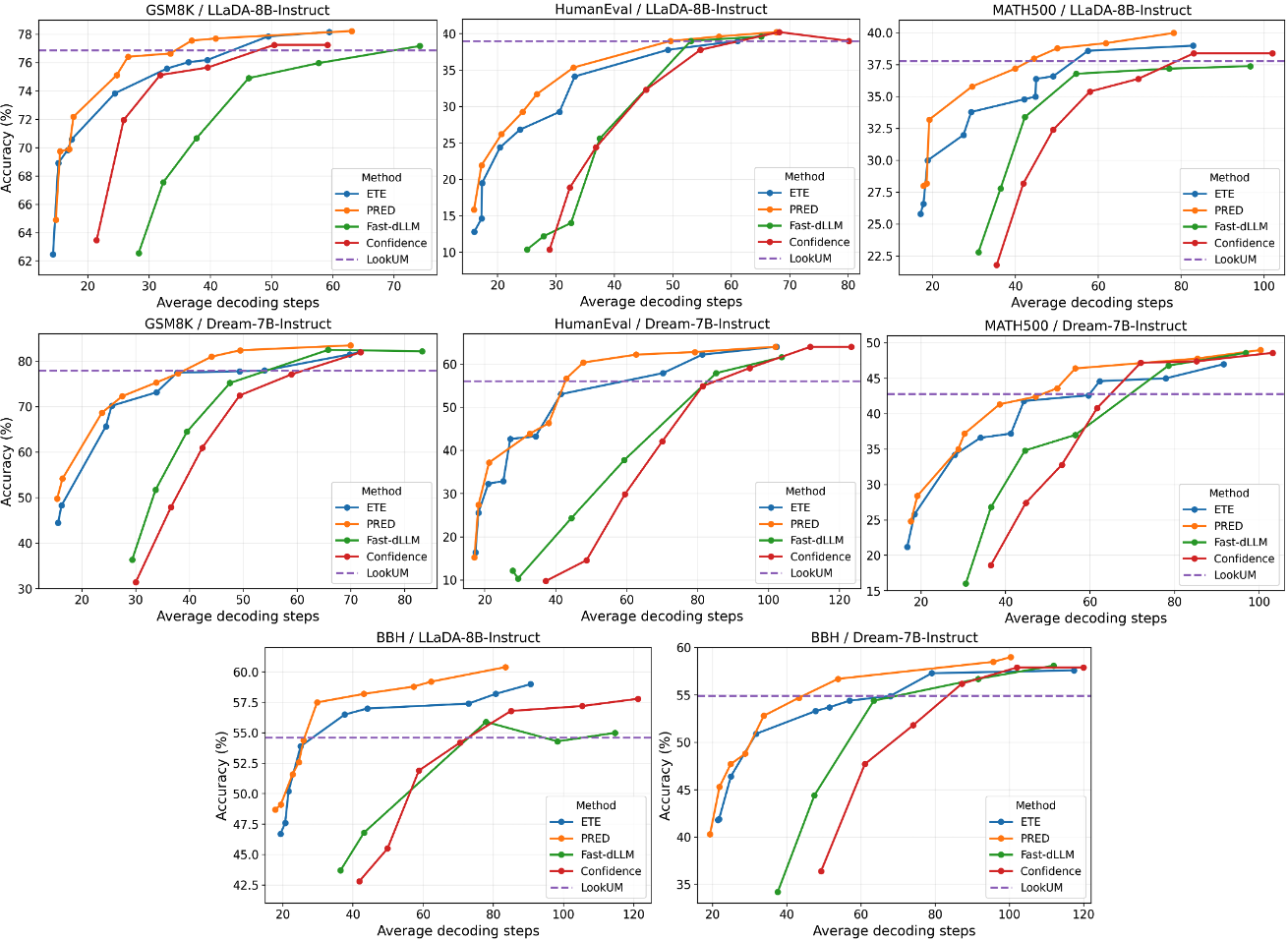}
    \vspace{-0.1in}
    \caption{\small Accuracy vs decoding steps across different models and benchmarks}
    \label{fig:results}
    \vspace{-0.15in}
\end{figure*}

We report the accuracy--efficiency trade-off under different decoding configurations in Figure~\ref{fig:results}. 
For PRED and ETE, we obtain different operating points by varying the \(N/C\) configurations, while for Conf and Fast-dLLM, we vary their confidence-related decoding thresholds.
Each point corresponds to one decoding configuration, and the curves show how accuracy changes with the decoding budget across different datasets. 
Here, decoding steps refer to the average number of model forward passes used during generation across evaluation examples. 
{Notably, LookUM inherently uses a fixed number of decoding steps and does not support varying the decoding budget. We therefore report its performance using the 128-step setting adopted in the original paper, shown as a horizontal dashed reference line in Figure~\ref{fig:results}.}
Based on Figure~\ref{fig:results}, we have the following observations.


First, search-based decoding methods ETE and PRED generally achieve better accuracy--efficiency trade-offs than purely confidence-based methods. Fast-dLLM and Conf achieve trade-offs comparable to those of the search-based methods only in a limited number of cases, such as MATH with Dream. This trend highlights the benefit of explicitly exploring alternative decoding trajectories beyond relying solely on confidence-aware token selection. {In addition, we notice that although LookUM uses a budget of 128 decoding steps, which is larger than the maximum budget of other methods, it fails to achieve the highest accuracy in most settings. This is likely related to its fixed-step sampling design, which is less flexible in allocating decoding computation than confidence-adaptive methods despite the use of a search mechanism.}

{Second, compared with ETE, the search-based baseline most closely related to PRED in terms of its decoding framework, PRED consistently achieves stronger performance. } For example, at around 40 decoding steps, PRED outperforms ETE by approximately 2.5 percentage points with LLaDA and about
5 percentage points with Dream. Since both methods explicitly search over alternative decoding trajectories, this improvement suggests
that planner-guided recursive exploration further strengthens search-based decoding by adaptively determining both when to search and how far to explore candidate trajectories. In Section~\ref{sec:abl}, we conduct ablation studies to verify that both the planner and multi-step exploration contribute to the overall performance improvements.


Third, the learned planner generalizes well across benchmarks: for each base model, the planner is trained only on the combined GSM8K and MATH training data and applied to all evaluation benchmarks without benchmark-specific retraining. Its consistent effectiveness on unseen benchmarks BBH and HumanEval suggests that the planner learns transferable search-triggering patterns rather than overfitting to the training tasks.

Finally, PRED also tends to achieve a higher maximum reachable accuracy across most settings. For example, on BBH with LLaDA, PRED achieves a best accuracy of 60.1\%, compared with 59.0\% for ETE, 55.9\% for Fast-dLLM, and 57.8\% for Conf. This suggests that adaptive multi-step exploration not only improves decoding efficiency but also helps the decoder identify higher-quality reasoning trajectories that may be difficult to reach with shallow search alone.

In addition, to provide a more comprehensive evaluation of decoding efficiency beyond the number of decoding steps, we analyze the per-step decoding latency of PRED and ETE in Appendix~\ref{appd:latency}. The results show that, while PRED requires additional batched forward passes during multi-step exploration, it generally increases the average per-step latency by only around 4\%, indicating that the practical overhead remains very small.

\vspace{-0.04in}
\subsection{Ablation Studies}\label{sec:abl}
\vspace{-0.03in}
We conduct two sets of ablation studies: component and design ablations, and an analysis of the effect of generation and block length. Due to space limitations, we present the latter in Appendix~\ref{sec:size}.


\vspace{-0.05in}
\subsubsection{Component and Design Ablations}\label{sec:nai}
\vspace{-0.05in}
In this subsection, we conduct component-wise ablations on MATH500 using LLaDA-8B-Instruct to show that both the learned search planner and the adaptive multi-step recursive exploration mechanism contribute to the overall performance gains.

Specifically, we vary the confidence threshold $C$ from 0.5 to 0.9 to obtain different accuracy--efficiency operating points. We compare four variants: (1) ETE, which performs one-step search with confidence-based search triggering; (2) ETE Multi-Step, which extends ETE with a fixed two-step linear rollout by advancing each candidate trajectory for one additional decoding step and selecting the candidate based on its accumulated two-step trajectory score; (3) full PRED; and (4) PRED without the learned planner, where the planner is replaced by the confidence-based triggering rule introduced in Section~\ref{sec:planner} and used for supervised warm-up. 

The four variants allow us to progressively examine the key design choices of PRED. As shown in Table~\ref{tab:ablation}, simply extending ETE from one-step to a fixed two-step rollout consistently degrades performance under comparable decoding budgets, indicating that increasing the rollout horizon alone is insufficient. In contrast, PRED without the planner generally outperforms ETE, demonstrating the benefit of adaptive multi-step and recursive exploration beyond fixed one-step search. Finally, full PRED consistently improves over the planner-free variant and achieves the highest accuracy across all threshold settings, confirming the contribution of the learned search planner.



\begin{table}[t]
\centering
\caption{\small Ablation results on MATH500 with LLaDA}
\vspace{-0.1in}
\label{tab:ablation}
\resizebox{0.5\textwidth}{!}{
\begin{tabular}{c|cc|cc|cc|cc}
\toprule
& \multicolumn{2}{c|}{ETE}
& \multicolumn{2}{c|}{ETE Multi-Step}
& \multicolumn{2}{c|}{PRED w/o Planner}
& \multicolumn{2}{c}{PRED} \\
$C$ & Steps & Acc. & Steps & Acc. & Steps & Acc. & Steps & Acc. \\
\midrule
0.5 & 17.10 & 25.8\% & 17.89 & 24.6\% & 17.64 & 26.8\% & 17.79 & 28.0\% \\
0.6 & 19.24 & 30.0\% & 20.04 & 29.2\% & 19.27 & 32.0\% & 19.21 & 33.2\% \\
0.7 & 29.34 & 33.8\% & 29.61 & 33.2\% & 29.58 & 34.6\% & 29.55 & 35.8\% \\
0.8 & 45.00 & 36.4\% & 43.92 & 34.2\% & 45.87 & 36.6\% & 44.52 & 38.0\% \\
0.9 & 82.87 & 39.0\% & 81.86 & 38.0\% & 79.38 & 39.4\% & 78.23 & 40.0\% \\
\bottomrule
\end{tabular}}
\vspace{-0.18in}
\end{table}

\vspace{-0.05in}
\section{Conclusion}
\vspace{-0.05in}

In this work, we present PRED, a planner-guided recursive exploration framework for efficient DLM decoding. PRED adaptively determines when to trigger additional search and how far candidate trajectories should be explored, combining a learned search planner with multi-step trajectory evaluation and recursive branch expansion. Experiments across multiple models and benchmarks demonstrate that PRED achieves a better accuracy--efficiency trade-off compared with confidence-based and existing search-based decoding methods.

\section*{Limitations}
While our method consistently improves the accuracy--efficiency trade-off in terms of the number of decoding steps, the adaptive multi-step exploration introduces slightly higher computational overhead per decoding step due to the batched forward passes required for hypothesis evaluation, even on high-end GPUs. We believe that this overhead will become increasingly negligible as hardware capabilities continue to advance and batched inference becomes more efficient.


\bibliography{custom}

\appendix

\section{Algorithm}
\label{appd:alg}
We present the detailed algorithm of PRED in Algorithm~\ref {alg:ours}.

\begin{algorithm*}[t]
\caption{Planner-Guided Recursive Exploration Algorithm}
\label{alg:ours}
\begin{algorithmic}[1]
\Require Model $p_\theta$, search planner $f_\phi$, current state $\mathbf{x}_t$, masked token set $\mathcal{M}$,
         beam width $k$, medium-confidence target $c^{\text{info}}$, position bias $\beta$,
         downstream threshold $C$, sample-quality weight $\alpha$,
         variance threshold $\tau$, rollout budget $T_{\text{max}}$

\State \textbf{// Planner-guided search triggering}
\State $\mathrm{r}_t \leftarrow \textsc{PlannerTrigger}(f_\phi, \mathbf{x}_t)$
\If{$\mathrm{r}_t = \texttt{False}$}
    \State \Return $\textsc{ConfidenceDecode}(p_\theta,\mathbf{x}_t)$
\Else

\State \textbf{// Step 1: Select $k$ exploration candidates}
\State Score each masked position $i \in \mathcal{M}$:
       $\quad\phi_i = -|c_i - c^{\text{info}}| + \beta\,(i - (b-1)\,n_b)$
\State $\mathcal{H} \leftarrow \mathrm{Top\text{-}}k\bigl(\{\phi_i\}_{i \in \mathcal{M}}\bigr)$

\State \textbf{// Step 2: Build $k$ hypotheses and run one-step rollout}
\For{each candidate $j \in \mathcal{H}$}
    \State $\mathbf{x}^j \leftarrow \mathbf{x}_t$ with position $j$ committed
\EndFor
\State $\{(\hat{x}^j, c^j)\}_{j\in\mathcal{H}} \leftarrow \textsc{BatchedForward}(p_\theta,\,\{\mathbf{x}^j\}_{j \in \mathcal{H}})$
\For{each $j \in \mathcal{H}$}
    \State \ $s_j \leftarrow \text{Eq.(\ref{eq:s})}$\hfill \Comment{Score $j$ by unlocked confidence mass}
\EndFor
\State \textbf{// Step 3: Variance gate — decide whether to run extra forward chain}
\If{$\operatorname{Var}(\{s_j\}_{j \in \mathcal{H}}) < \tau$}
    \For{$r = 1, \ldots, T_{\text{max}}-1$}
        \State $\{x^j\} \leftarrow \textsc{Commit}(\{(\hat{x}^j, c^j)\})$ for all $j \in \mathcal{H}$
        \State $\{(\hat{x}^j, c^j)\} \leftarrow \textsc{BatchedForward}(p_\theta,\,\{\mathbf{x}^j\}_{j \in \mathcal{H}})$
        \State $\mathrm{r}_j \leftarrow \textsc{PlannerTrigger}(f_\phi, \hat{x}^j, c^j)$ for all $j \in \mathcal{H}$

\If{\textbf{all} $\mathrm{r}_j = \text{True}$} \hfill\Comment{Case 2: all trigger}
    \State $j^* \leftarrow \arg\max_j\,(s_j + \log \bar{c}^j_\mathcal{F})$
    \State \Return $(\hat{x}^{j^*}, c^{j^*})$ as initial state for next decoding round
\ElsIf{\textbf{no} $\mathrm{r}_j = \text{True}$} \hfill\Comment{Case 1: no trigger}
    \State $s_j \leftarrow \text{Eq.(\ref{eq:s})}$ \quad for all $j \in \mathcal{H}$
    \If{$\operatorname{Var}(s_j) \geq \tau$}
        \State $j^* \leftarrow \arg\max_j\, {s}_j$;\ \ \textbf{go to} Commit
    \EndIf
\Else \hfill\Comment{\textit{Case 3: Mixed — shrink to stable hypotheses}}
    \State $\mathcal{H} \leftarrow \{j \in \mathcal{H} : \mathrm{r}_j = \texttt{False}\}$
    \State $s_j \leftarrow \text{Eq.(\ref{eq:s})}$ \quad for all $j \in \mathcal{H}$
    \If{$|\mathcal{H}| = 1$}
        \State $j^* \leftarrow$ the unique element in $\mathcal{H}$; \textbf{go to} Commit
    \EndIf
    \If{$\operatorname{Var}(s_j) \geq \tau$}
        \State $j^* \leftarrow \arg\max_j\, {s}_j$;\ \ \textbf{go to} Commit
    \EndIf
\EndIf
        
       \quad $j^* \leftarrow \arg\max_j\, {s}_j$;\ \ \textbf{go to} Commit
    \EndFor
\Else
\quad $j^* \leftarrow \arg\max_j\, {s}_j$;\ \ \textbf{go to} Commit
\EndIf

\State \textbf{// Final Step: Commit}

\State $\mathbf{x}^{j^*} \leftarrow \textsc{Commit}(\hat{x}^{j^*}, c^j)$
\State \Return $\mathbf{x}^{j^*}$

\EndIf
\end{algorithmic}
\end{algorithm*}

\section{Additional Implementation Details}
\label{appd:training}

\paragraph{Planner Architecture and Supervised Warm-up.}
For both base DLMs, we use the same planner architecture while adapting the input hidden-state dimension to the corresponding backbone. Specifically, the planner uses a lightweight Transformer head with a projection dimension of 64, hidden dimension of 64, 4 attention heads, and an FFN dimension of 256. We use a dropout rate of 0.1 and a maximum block length of 128 for the learnable positional embeddings. The hidden-state input dimension is 4096 for LLaDA-8B-Instruct and 3584 for Dream-v0-Instruct-7B. 

During supervised warm-up, we optimize the planner using binary cross-entropy with logits. We use AdamW with a learning rate of $3\times10^{-4}$ and weight decay of 0.01, with a batch size of 64. Training is run for at most 20 epochs with early stopping based on validation loss and a patience of 3 epochs. We use a random seed of 41 for all runs.

\paragraph{GRPO Refinement.}
After supervised warm-up, we further refine the planner using on-policy group-relative policy optimization. Training is performed on 4 GPUs with distributed data parallelism. For each prompt, we sample a group of $G=16$ trajectories, with 4 prompts used per update, resulting in 64 trajectories per optimization step. We use AdamW with a learning rate of $1\times10^{-4}$, weight decay of 0.01, and gradient clipping with a maximum norm of 1.0. We reserve 5\% of the training split for validation and evaluate the planner every 50 updates.

For the trajectory-level reward, a correct generation receives an accuracy reward of $\lambda_{\mathrm{acc}}=2.0$, while inference cost is measured using the number of model forward evaluations (NFEs) with cost weight $\lambda_{\mathrm{cost}}=0.01$. We normalize the group-relative advantage using a fixed scale,
\[
A_i = \frac{R_i-\bar{R}}{2.0},
\]
without additional advantage clipping. During policy sampling, we use $\epsilon=0.05$ to maintain exploration. We further apply KL regularization with coefficient 0.01 against the frozen planner obtained after supervised warm-up, while the entropy coefficient is set to 0. We use a random seed of 41 throughout GRPO training.

\noindent\textbf{Decoding Hyperparameter Configurations.}
For ETE and PRED, we vary the per-block unlocking budget $N$ over powers of two, $N\in\{1,2,4,8,16\}$, and pair each value with confidence thresholds $C$ selected at increments of 0.05 according to the calibration rule in Section~\ref{sec:hyperparameter}. These $N/C$ combinations are used to obtain the accuracy--efficiency operating points reported in our experiments.

\section{Justification for the log-scale calibration rule}\label{appd:theo}
\begin{figure}[h]
    \centering
    \includegraphics[width=1\linewidth]{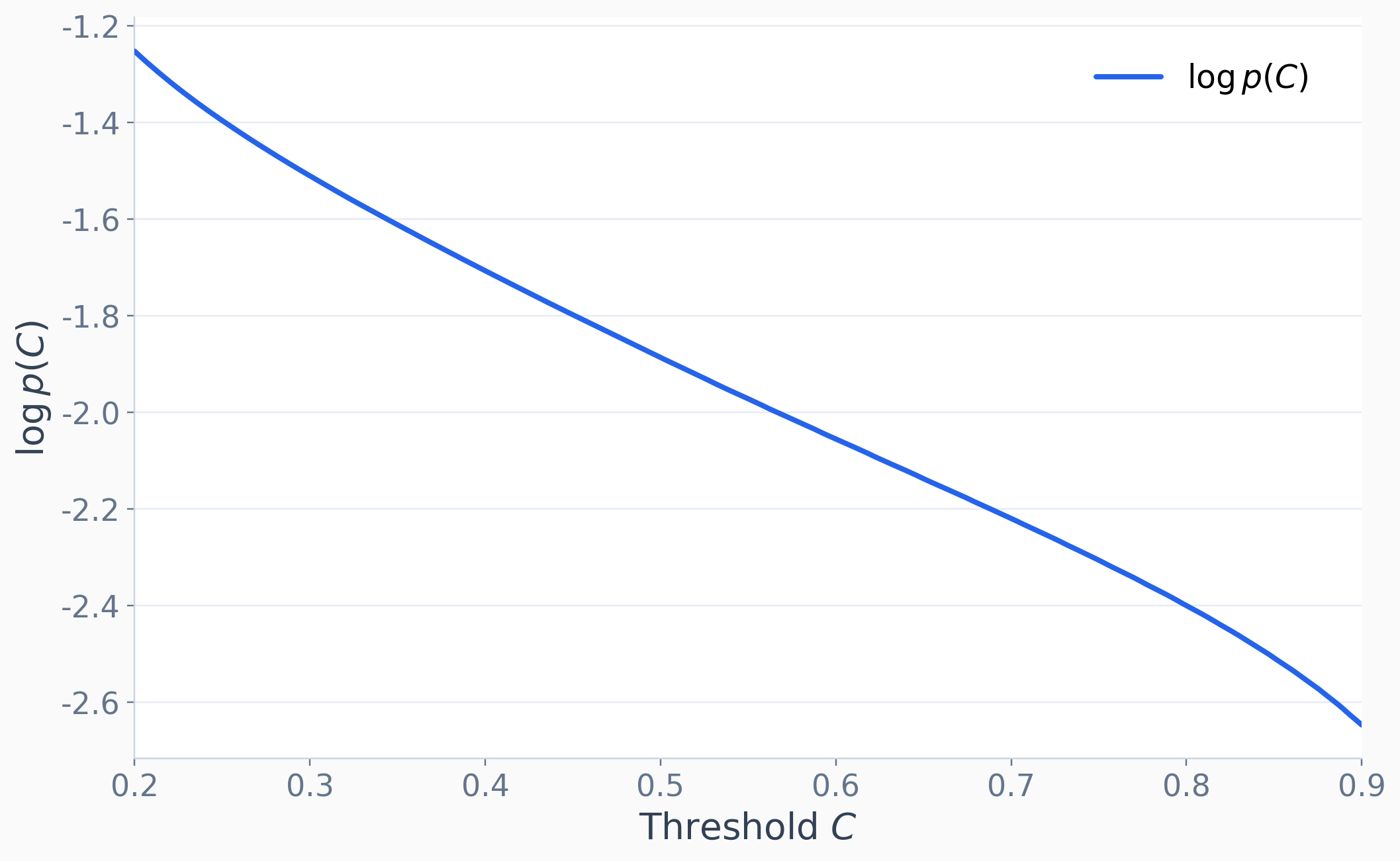}
    \caption{ Relationship between $\log(p(c))$ and $C$ (evaluated with MATH500).}
    \label{fig:log}
\end{figure}
We provide an additional justification for why the preferred confidence threshold \(C\) increases approximately logarithmically with the per-block unlocking budget \(N\).

For a fixed confidence threshold \(C\), the expected fraction of tokens that can be committed in one refinement step is
\[
p(C) = \Pr(c \ge C).
\]
After \(N\) rounds the expected fraction of tokens committed within the block can be approximated as $1-(1-p(C))^N$.
If we want to maintain a comparable block progression rate with different values of $N$, this probability should remain roughly at a constant level, then we have
\[
(1-p(C))^N \approx \text{const}.
\]
Taking log for both sides:
\[
N \log(1-p(C))=-\text{const}.
\]
When $p$ is small, we have $\log(1-p)\approx -p$, then we have
\begin{equation}\label{eq:cons}
    p(C) \approx \frac{\text{const}}{N} 
\end{equation}

Thus, as \(N\) increases, the required per-step commitment probability \(p(C)\) decreases, meaning that a larger confidence threshold \(C\) can be used.

To connect this relationship to the logarithmic schedule, we empirically examine the tail probability of masked-token confidence. 
As shown in Figure~\ref{fig:log}, over the practical calibration range, the log-tail probability \(\log \Pr(c \ge C)\) decreases approximately linearly with \(C\) when $C \in [0.2, 0.9]$. 
This suggests that the high-confidence tail can be approximated by an exponential form:
\[
p(C) \approx \exp(-\lambda C),
\]
for some \(\lambda > 0\). 
Combining this empirical approximation with Eq.(\ref{eq:cons}) yield:
\[
\exp(-\lambda C) \approx \frac{\text{const}}{N} .
\]
Therefore,
\[
C \approx \frac{1}{\lambda}\log N + \mathrm{const}.
\]
This provides a theoretical justification for using a log-scale guideline to select \(C\) as the unlocking budget \(N\) varies, while the exact threshold is chosen empirically on the calibration set.

\section{Additional Empirical Results}
\subsection{Analysis on Latency}\label{appd:latency}
In Section~\ref{sec:main}, we use the average number of forward steps per example to evaluate decoding efficiency. This metric provides a hardware-independent estimate of inference cost. 
However, our adaptive multi-step exploration method may require more batched forward passes during rollout, since multiple candidate trajectories are evaluated in parallel within a single rollout step. 
As a result, this additional batched computation may increase the overall generation latency, even under a similar number of decoding steps.
In Section~\ref{sec:main}, we use the average number of forward steps per example as a hardware-independent measure of decoding efficiency. Beyond the number of decoding steps, PRED may introduce additional per-step latency compared with one-step search, since its adaptive multi-step exploration can require additional batched forward passes to evaluate candidate trajectories. To quantify this additional search overhead, we further examine the practical latency introduced by PRED relative to ETE, which provides a directly comparable one-step search baseline.

We measure generation latency under beam size $k=4$, which corresponds to the batch size used for parallel candidate evaluation. 
Table~\ref{tab:beam} reports the average time per decoding step on H200 and B200 GPUs, measured in milliseconds. We report both PRED with and without the learned planner. For PRED without the planner, we replace the learned planner with the confidence-based triggering rule described in Section~\ref{sec:planner}, which is also used to initialize the planner during supervised warm-up.
Notably, this value is not computed by timing an isolated model forward pass. 
Instead, we measure the end-to-end generation time, including model forward passes, score computation, candidate selection, branch evaluation, and other decoding overhead, and divide it by the average number of forward steps. 
Thus, the reported value should be interpreted as the average time corresponding to one decoding step reported in Figure~\ref{fig:results}.

Results in Table~\ref{tab:beam} show that, due to the additional rollout and branch expansion operations, PRED introduces slightly higher latency than ETE. However, the increase remains small: under the beam size $k=4$ used in our experiments, PRED increases the per-step latency by only around $4\%$ on both H200 and B200 GPUs. 
This is because all candidate branches at each rollout step are processed together in a single batched forward pass, so the additional parallel computation introduces only limited latency overhead, especially on high-performance GPUs.

Moreover, PRED with and without the learned planner exhibit nearly identical latency, indicating that the planner itself introduces negligible computational overhead. These results suggest that the additional computation introduced by PRED remains modest in practice.

\begin{table}[t]
\centering
\caption{Average latency per decoding step (ms)}
\vspace{-0.1in}\label{tab:beam}
\resizebox{0.3\textwidth}{!}{
\begin{tabular}{l|cc}
\toprule
Method & H200 & B200 \\
\midrule
ETE              & 33.46 & 21.65 \\
PRED w/ planner  & 34.90 & 22.58 \\
PRED w/o planner & 34.85 & 22.65 \\
\bottomrule
\end{tabular}}
\label{tab:latency}
\vspace{-0.14in}
\end{table}

\subsection{Additional Ablation Studies: Effect of Generation Length / Block Size}\label{sec:size}

In this ablation study, we evaluate PRED under different generation-length and block-size configurations to study whether its effectiveness can generalize well to other decoding setups.

\begin{figure}
    \centering
    \vspace{-0.05in}
    \includegraphics[width=0.92\linewidth]{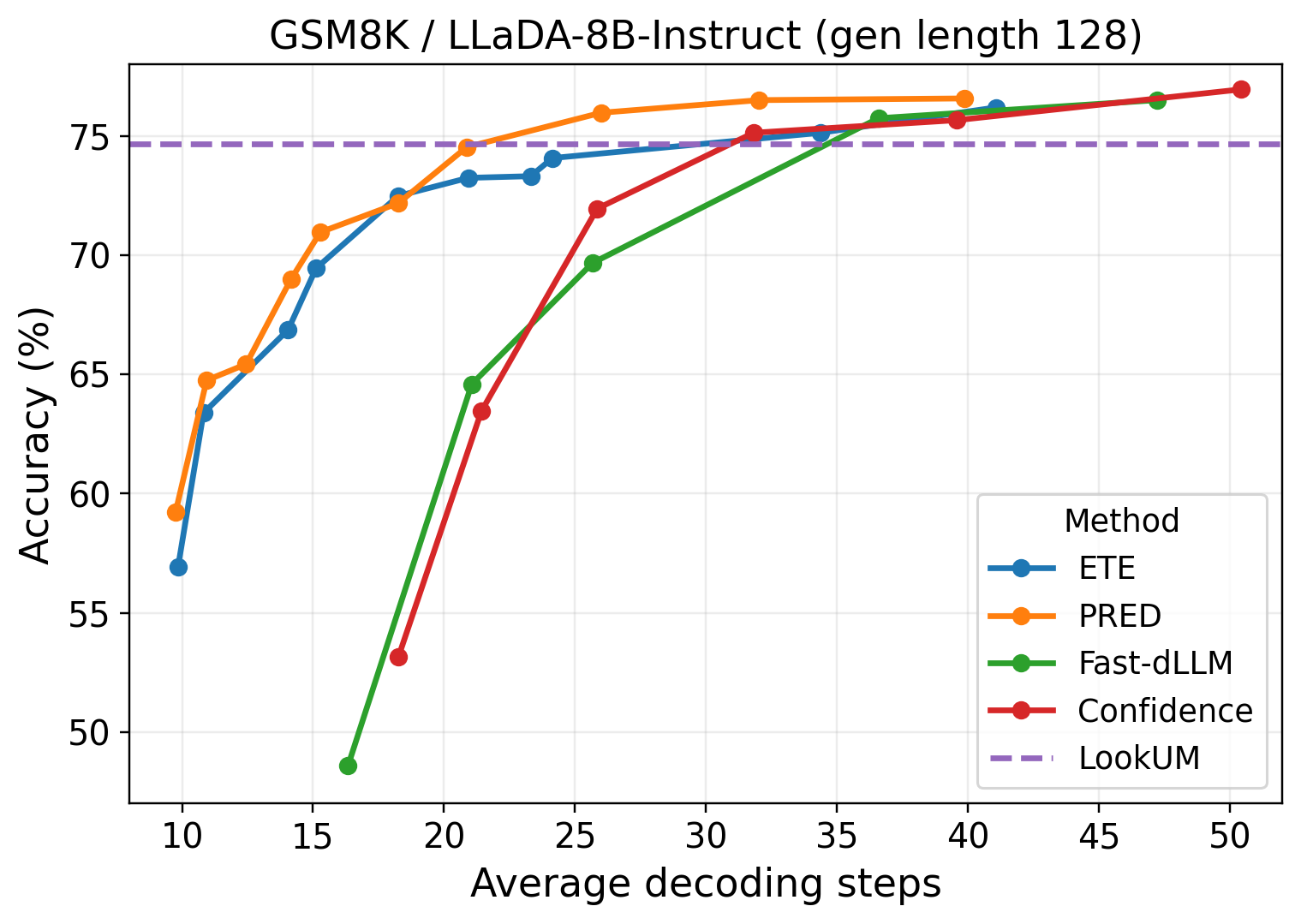}
    \vspace{-0.1in}
   \caption{\small Results on GSM8K with LLaDA-8B-Instruct under a generation length of 128 and block size of 32.}
   \vspace{0.05in}
    \label{fig:128}
    \centering
    \includegraphics[width=0.92\linewidth]{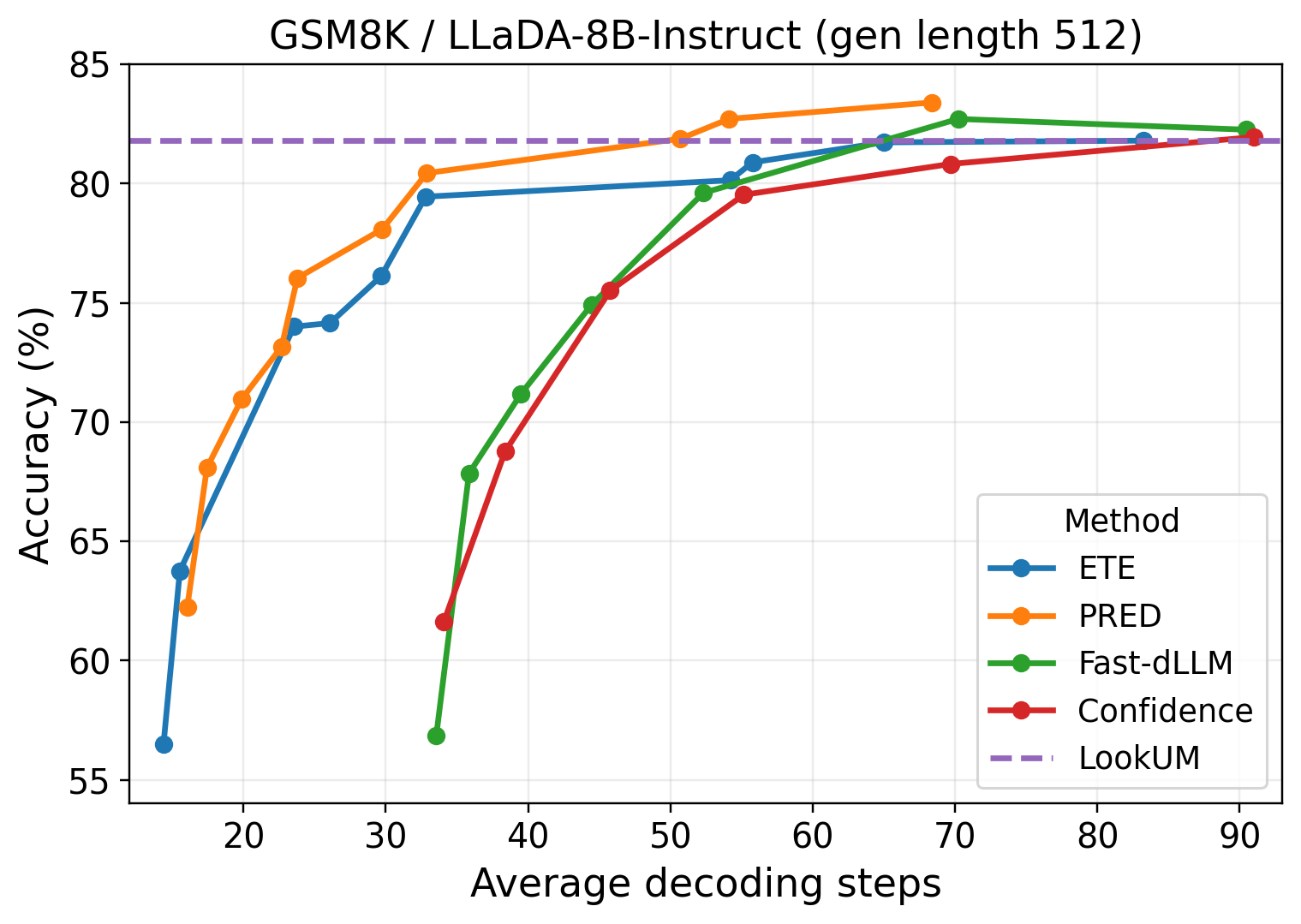}
    \vspace{-0.07in}
    \caption{\small Results on GSM8K with LLaDA-8B-Instruct under a generation length of 512 and block size of 64.}
    \label{fig:512}
    \vspace{-0.17in}
\end{figure}

Figures~\ref{fig:128} and~\ref{fig:512} report the results on GSM8K under generation-length/block-size settings of $128/32$ and $512/64$, respectively. From the results, we can observe that PRED consistently maintains a better accuracy--decoding steps trade-off under both decoding configurations, suggesting that its effectiveness generalizes well to different generation lengths and block sizes.

\section{Disclosure of AI Assistant Usage}

AI assistants were used solely for language polishing and proofreading during manuscript preparation. 
Specifically, they were used to improve grammar, spelling, clarity, and overall readability of the text. 
All technical content, experiments, analyses, and conclusions were developed and verified by the authors.

\section{Artifacts, Licensing, and Intended Use} 
All datasets and pretrained models used in this work are publicly available and are utilized in accordance with their respective licenses. Our use is strictly limited to research and evaluation purposes, consistent with the intended scope of each artifact. No third-party artifacts are redistributed or modified in terms of their access conditions.
\end{document}